\documentclass[10pt,twocolumn,letterpaper]{article}

\usepackage{cvpr}              % To produce the CAMERA-READY version
\usepackage{multirow}
\usepackage{multicol}
\usepackage{graphicx}
\usepackage{amsmath}
\usepackage{amssymb}
\usepackage{xcolor}
\usepackage{pifont}
\usepackage[font=small,labelfont=bf]{caption}
\definecolor{cvprblue}{rgb}{0.21,0.49,0.74}
\usepackage[pagebackref,breaklinks,colorlinks,citecolor=cvprblue]{hyperref}

\definecolor{maroon}{cmyk}{0,0.87,0.68,0.32}
\definecolor{customgreen}{cmyk}{1,0,1,0.5}
\newcommand{\cmark}{\color{customgreen}\ding{51}}%
\newcommand{\xmark}{\color{maroon}\ding{55}}%

\newcommand{\Ours}{Diffusion-HPC}

\def\paperID{265} % *** Enter the Paper ID here
\def\confName{3DV\xspace}
\def\confYear{2024\xspace}

\title{Diffusion-HPC: Synthetic Data Generation for Human Mesh Recovery in Challenging Domains}

\author{Zhenzhen Weng, Laura Bravo-Sánchez, Serena Yeung-Levy \\
Stanford University \\
{\tt\small \{zzweng, lmbravo, syyeung\}@stanford.edu }
}

\begin{document}
\twocolumn[{%
\renewcommand\twocolumn[1][]{#1}%
\maketitle
\begin{center}
    \centering
    \captionsetup{type=figure}
    \includegraphics[width=\textwidth]{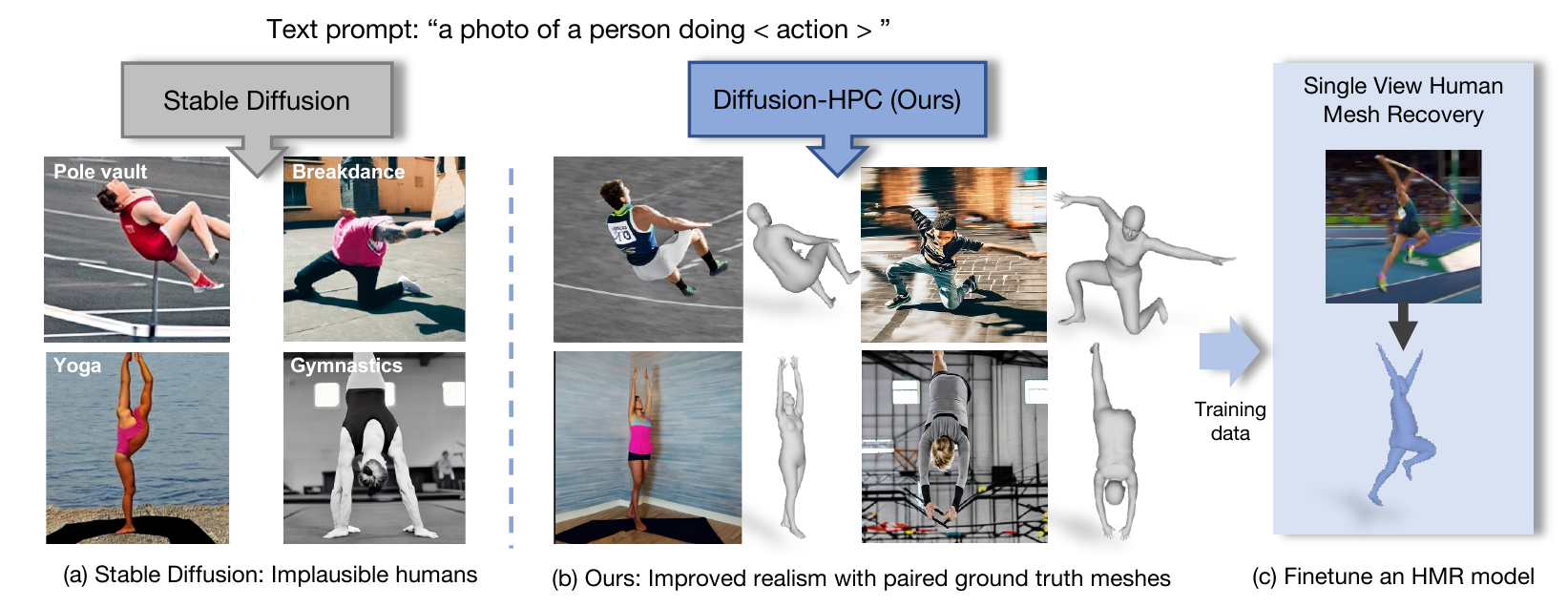}
    \captionof{figure}{
    We propose Diffusion model with Human Pose Correction (\Ours{}), a synthetic image generation strategy with paired with ground-truth meshes to improve the performance of Human Mesh Recovery (HMR) models on domains with challenging poses and/or limited data. \Ours{} is a text-conditioned method that addresses the implausibility of human generations from Stable Diffusion \cite{Rombach_2022_CVPR}, a large pre-trained text-conditioned generative model, while preserving the inherent flexibility of such models.
    % Our synthetic image generation strategy \Ours{} addresses the limitations of diffusion models by injecting human pose
    % \SY{sounds like your method requires a pose as an input, which is weaker. It leverages pose prior in the generation process but can work w/o changing the exact same text-conditioned setup as SD} 
    }
    \label{fig:pull_figure}
\end{center}%
}]
\maketitle
\begin{abstract}
Recent text-to-image generative models have exhibited remarkable abilities in generating high-fidelity and photo-realistic images. However, despite the visually impressive results, these models often struggle to preserve plausible human structure in the generations. Due to this reason, while generative models have shown promising results in aiding downstream image recognition tasks by generating large volumes of synthetic data, 
they are not suitable for improving downstream human pose perception and understanding. In this work, we propose a Diffusion model with Human Pose Correction (\Ours{}), a text-conditioned method that generates photo-realistic images with plausible posed humans by injecting prior knowledge about human body structure. Our generated images are accompanied by 3D meshes that serve as ground truths for improving Human Mesh Recovery tasks, where a shortage of 3D training data has long been an issue. Furthermore, we show that \Ours{} effectively improves the realism of human generations under varying conditioning strategies.
% We show that \Ours{} effectively improves the realism of human generations. Furthermore, as the generations are accompanied by 3D meshes that serve as ground truths, \Ours{}'s generated image-mesh pairs are well-suited for downstream human mesh recovery task, where a shortage of 3D training data has long been an issue.
\footnote{Code: \url{https://github.com/ZZWENG/Diffusion_HPC}}
\end{abstract}    
\vspace{-0.7cm}
\section{Introduction}
% Understanding the world from 2D is an ill-posed problem(?) But extracting 3D is costly in terms of data, equipment, incompatible with some aspects of in-the-wild human life. Single view human mesh recovery to the rescue.

In recent years, large-scale text-conditioned image generation models such as GLIDE \cite{nichol2021glide}, Imagen \cite{saharia2022photorealistic} and Stable Diffusion \cite{Rombach_2022_CVPR} have impressed the research community with their exceptional generative and compositional capabilities, owing to their training on extremely large image-text datasets \cite{schuhmann2022laion} and use of advanced model architectures \cite{ho2020denoising,dhariwal2021diffusion}. Not only do these generative models drastically elevate the quality and efficiency of content creation, but they also exhibit promising potential for enhancing other visual tasks. As shown in \citet{he2022synthetic}, large text-conditioned generative models such as GLIDE \cite{nichol2021glide} are able to generate high-quality images targeted for a specific label space (i.e. domain customization), thus making it an ideal choice for synthetic data generation to aid in downstream image recognition tasks such as single-view Human Mesh Recovery (HMR) \cite{kolotouros2019learning} where securing annotations can be not only costly, but incompatible in the wild.

% \SY{maybe list the specific tasks, to contrast with yours}

Despite the benefits of synthetic data for image recognition tasks, these text-conditioned generative models have thus far lacked the capability of advancing human pose understanding tasks.
% \SY{is it just downstream tasks? You also help SD generate better images without considering downstream tasks} 
This is because these models do not explicitly model the underlying structure of human bodies and thus frequently encounter difficulties in preserving realistic human anatomy in their generated outputs. As Figure \ref{fig:pull_figure} (a) shows, generating realistic human poses embedded in plausible scenes is a known
% \SY{is it known/reported that you can cite? Or just clear from your work}
limitation \cite{hugging_face_2022} of generative diffusion models such as Stable Diffusion \cite{Rombach_2022_CVPR}.

% Understanding the subtleties of human movement and behaviour from images is one of the basic ideas in computer vision. In recent years, single-view Human Mesh Recovery (HMR) \cite{kolotouros2019learning} has received increasing attention as an alternative for deriving meaningful information from a 2D representation of a 3D world. Despite being capable of adapting to domains where only 2D ground truth or pseudo-ground truth is available. Most HMR methods continue to rely on 3D ground truth for successfully modeling challenging human poses. However, securing 3D annotations can be not only costly, but incompatible for some of these domains.

% Recent works such as \citet{he2022synthetic}, have studied how synthetic images generated with large pre-trained diffusion models can bridge the data gap in image classification tasks. Yet, as Figure \ref{fig:pull_figure} shows generating realistic human poses embedded in plausible scenes is a known limitation of diffusion models. 

In this work, we present Diffusion model with Human Pose Correction (\Ours{}), a method that addresses the implausibility of human generations from large pre-trained text-conditioned generative models. Our intuition is that we can rectify the generated unrealistic humans (e.g. with additional limbs in non-anatomical locations) by integrating stronger human pose priors within the generation process.
% \SY{remaining part of this paragraph a little unclear. Unclear if it is a variant of your method, or if there are just simultaneous other benefits} 
Thereby, we extend the capability of pre-trained diffusion models, such as Stable Diffusion, to produce a large variety of synthetic scenes for a target domain with minimal user input. Further, unlike base diffusion models our approach produces pairs of images and ground truth meshes as a result of including body pose priors in the generation process.
% \SY{unclear how, clarify at least a high-level idea} 
These image-mesh pairs can then be employed to improve existing single-view Human Mesh Recovery methods on challenging data-scarce domains (See Figure \ref{fig:pull_figure}b, c). %We further show the flexibility of \Ours{} by validating the generation potential when including source images as additional user input.
%a domain in question with minimal user input and their learned understanding of the underlying human pose distributions. To the best of our knowledge, our work presents the first training-free attempt that addresses the challenges in generating realistic humans using large pre-trained generative models.  
% Moreover, aided by the pairs of generated scenes and meshes we show that our additional data can be employed to finetune existing single-view HMR methods to improve their performance on challenging domains. % See figure 1b,c
% We further showcase the utility of our generation method in bootstrapping training data for downstream human mesh recovery task on challenging domains through datasets SMART \cite{chen2021sportscap} and Ski-Pose dataset \cite{rhodin2018learning,sporri2016reasearch}.
% across a variety of competitive sports categories with challenging poses where we obtain an improvement of \Laura{\#} over existing methods. We also demonstrate the efficacy of our strategy under fast and extreme in-the-wild motions on the Ski-Pose dataset \cite{rhodin2018learning,sporri2016reasearch}, and show experiments under low-data regimes.
In summary, we make the following contributions.
\begin{itemize}[noitemsep,topsep=0pt]
    \item Motivated by the implausible humans produced by diffusion models, we propose a simple and effective method \Ours{} to rectify the implausibility of human generations that often occur in Stable Diffusion \cite{Rombach_2022_CVPR} results. To the best of our knowledge, our work presents the first training-free method that addresses the challenges in generating realistic humans by injecting human body structure priors within the generation process.
    % using large pre-trained generative models. \SY{your use of the word `` using'' sounds like your key insight is to improve SD by using large pretrained models. But really your key insight is injecting the structure to improve.}
    \item %We broaden the range of downstream utilities enabled by state-of-the-art generative models 
    We show that the synthetic images with corresponding 3D ground truth produced by our method are capable of adapting Human Mesh Recovery models to challenging domains (e.g. competitive sports) where supervision is limited and hard to obtain. Models finetuned with \Ours{}'s synthetic data achieve 2.6\% PCK and 4.6 PA-MPJPE improvement on SMART \cite{chen2021sportscap} and Ski-Pose 3D \cite{rhodin2018learning,sporri2016reasearch}, respectively.
    \item We quantitatively validate the improved quality of our generated images over existing text-to-image as well as state-of-the-art pose-to-image generative models.
    % \SY{be clear that SD cannot do this. Also, mention your performance.}
\end{itemize}
% We will make our code publicly available.

\section{Related Work}
\subsection{Using synthetic data to improve HMR}
Previous works \cite{he2022synthetic} have recognized the capability of state-of-the-art text-conditioned generative models \cite{nichol2021glide} for generating training data for downstream image recognition tasks. However, the poor quality of the generated person images effectively precludes the extension of this capacity to tasks such as 3D human pose understanding (e.g. human mesh recovery).
% \SY{is this verified or citable?}
Due to the challenge in collecting 3D ground truths for end-to-end training of human mesh recovery models, many previous works have considered leveraging synthetic data. Typically, these works create 2D renderings of 3D posed human models from graphics engines \cite{varol2017learning,doersch2019sim2real,patel2021agora}, with \citet{Black_CVPR_2023} being the most comprehensive effort. However, this approach possesses multiple disadvantages. First, 
% synthetic data generated from 
% manually defined pipelines may not accurately reflect the broad range of poses that exist in real-world data 
% \SY{somewhat unclear... Are they all really manual? In what way does it not accurately reflect real world data? Are you trying to contrast with your work / does your method clearly reflect real world data?} 
the variety of the generated poses is limited by the pose data source. Second, a large and diverse training set is needed to cover all possible poses of interest, which makes storing and sharing such data costly and inefficient. To address these, recent work \citet{weng2022domain} proposes a data-efficient way by rendering SMPL bodies with poses sampled from the estimated pose distributions from real data, but since the body textures are predicted and warped from real images, the renderings are not photo-realistic. Analogously, \citet{STRAPS2020BMVC} generates synthetic data online to improve diverse body shape estimation.

In contrast, using conditional generative models \citep{ho2020denoising,Rombach_2022_CVPR} such as ours to synthesize data has a few advantages. First, large generative models can produce high-fidelity photo-realistic images closer to real data since they are trained on internet-scale real-world data (e.g. LAION-5B \cite{schuhmann2022laion}). Second, they allow easy control of the generation style via detailed prompting, and stochasticity in the generation process results in more diverse and potentially unlimited synthetic data. However, although there have been some attempts to explore the use of generative models for image classification and object detection \cite{zhang2021datasetgan,he2022synthetic}, their usage in human mesh recovery has not yet been investigated due to the poor quality of the generated person images. \Ours{} is the first approach that uses conditional generative models to produce synthetic data that are useful for human mesh recovery, broadening the range of downstream utilities of SoTA generative models.
% \SY{I don't think the point is you are the first to use synthetic images generated by Diffusion-HPC, since this is a new method, are you trying to say you are the first to use diffusion models at all or something along those lines} 

\subsection{Conditional generation of posed humans}
Recently, there has been a growing focus on conditional generation of posed humans in the form of images/videos \cite{zhang2023adding,hu2023animate,wang2023disco,xu2023magicanimate}, body models \cite{guo2022generating,delmas2022posescript} or NeRF \cite{dong2023ag3d,weng2023zeroavatar,cao2023dreamavatar}. For synthesizing posed humans, most works focus either on text-conditioned or pose-conditioned generation. 
In terms of text-conditioned approaches state-of-the-art general image generation models such as Stable Diffusion \cite{Rombach_2022_CVPR} have shown impressive capability in producing high-resolution and realistic images. But as a known limitation \cite{hugging_face_2022}, they frequently struggle to preserve the correct anatomy of human bodies. An alternate line of research focuses directly on text-conditioned human pose or motion generation \cite{guo2022generating,delmas2022posescript}. These generative models are trained on large 3D human motion database \cite{mahmood2019amass} with paired textual descriptions. But since they output parameters of human body models \cite{loper2015smpl}, they bypass the issue of preserving anatomical structure. However, since human motion databases do not come with paired RGB data, these works are unable to produce human textures and background.

On the other hand, previous works have explored generating images of full-body humans conditioning on body pose \cite{albahar2021pose,knoche2020reposing,men2020controllable,brooks2022hallucinating}. \citet{albahar2021pose,knoche2020reposing,men2020controllable} consider the task of ``reposing", where the goal is to synthesize images of people in a novel pose, based on a reference image of that person and the new pose. More recently and relevant to our work, Brooks \etal \citep{brooks2022hallucinating} proposed a pose-conditioned image synthesis model that dispenses with the reference images by generating reasonable backgrounds. In contrast to the above works, our proposed \Ours{} is a person image synthesis method flexible enough to allow for both text and pose conditioned generation and does not require additional training or explicit pose annotations from a target domain to produce diverse humans and scenes. Another closely related work is ControlNet \cite{zhang2023adding}, where posed-conditioned images are obtained from generative models, yet unlike our method ControlNet requires finetuning on large amounts of real paired data (i.e. 2D keypoints, images and captions).

% \SY{emphasize more the big picture of your contribution, and why you have advantages / how exciting your work is}
% Explain how the hallucinating paper is the closest work we can compare to and how its approach is actually very impractical/different for our HMR purposes.
%
%rather than reposing we have pose-guided generation where we don't just repose and thus more diverse humans and scenes\

\subsection{Editing \& composing large pre-trained models}
% \vspace{-0.1cm}
% Look at the "Composing pre-trained models. " section under their related work. https://arxiv.org/pdf/2210.11522.pdf
Foundation models (e.g., Imagen \cite{saharia2022photorealistic}, Stable Diffusion \cite{Rombach_2022_CVPR}) that are trained on large amounts of broad data have demonstrated impressive generative and few-shot learning capabilities across a wide spectrum of tasks. Additionally, the scale of information they have seen and learned, allows these models to be adapted to further downstream tasks. For these reasons, editing or composing large pre-trained models has been widely studied recently. Among these works the most closely related to our approach are the training-free methods that utilize pre-trained diffusion models to perform global or local image editing \cite{hertz2022prompt,parmar2023zero,avrahami2022blended,avrahami2022latent} (e.g. inpainting, style transfer, etc.). They are ``training-free" in the sense that editing is done by injecting knowledge into the denoising process during inference and therefore no additional model finetuning is needed. Analogously, our method \Ours{} improves plausibility of human generations by injecting human body priors in the form of posed SMPL \cite{loper2015smpl} body models. 
% To the best of our knowledge, ours is the first method that aims to improve the human pose realism of pre-trained diffusion models.

% \citet{avrahami2022blended, avrahami2022blended} perform text-guided local editing in the pixel and latent space of pre-trained diffusion models to blend a novel object into a specific region of an image.

\begin{figure*}[t]
\centering
\includegraphics[width=\textwidth]{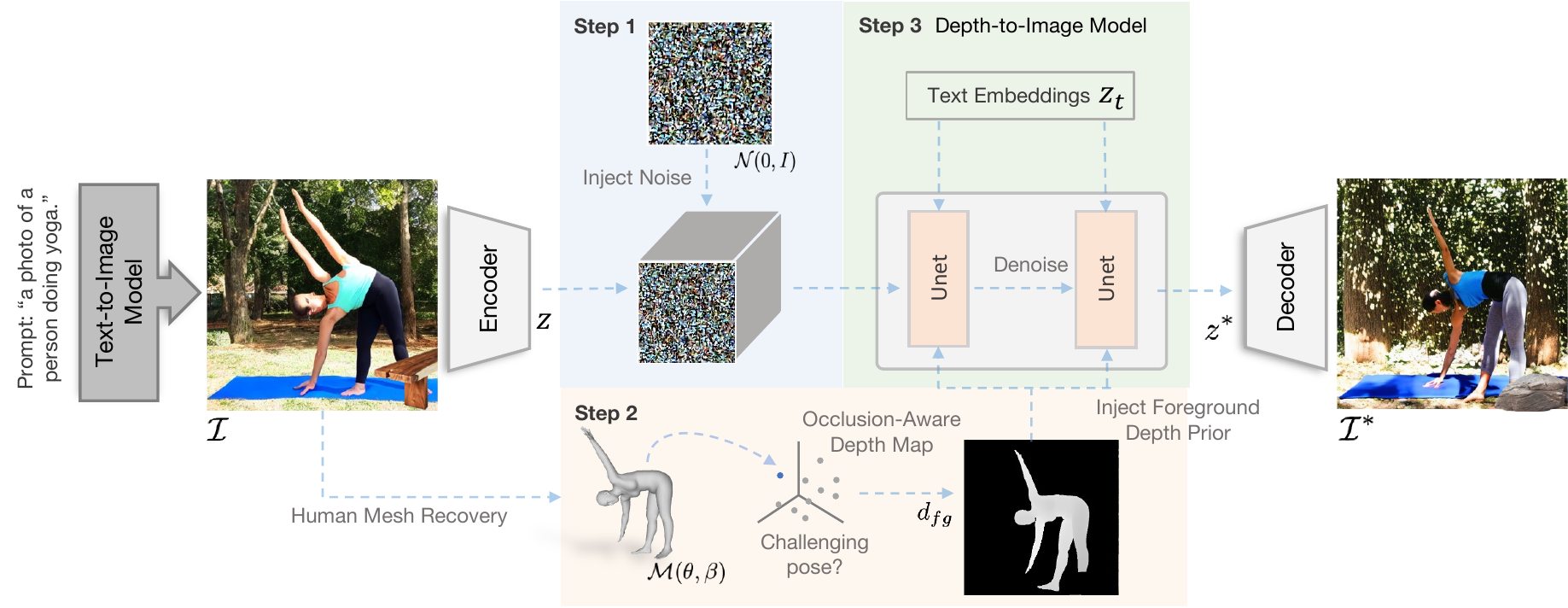}
\caption{Overview of \Ours{}. The generation process can be broken down into 3 steps. \textcolor{cyan}{Step 1}: Obtaining image latents $z$ from the initial generation $\mathcal{I}$ of a pre-trained text-to-image model (i.e. Stable Diffusion \cite{Rombach_2022_CVPR}) and injecting noise. \textcolor{pink}{Step 2}: Estimating human body mesh $\mathcal{M} (\theta, \beta)$ from $\mathcal{I}$. If the pose is challenging based on a pose prior (i.e. VPoser \cite{SMPL-X:2019}) then render the mesh's depth map $d_{fg}$ and introduce occlusions via object masks obtained from a segmentation model. \textcolor{teal}{Step 3}: Using the latents $z$, foreground depths, and the text embeddings $t$ as guide for the final generation $\mathcal{I^*}$. 
% \SY{ideally connect steps better to the main text. in main text, refer to these. maybe name them and label them for each colored box in the figure? Since there's a lot going on in the figure. Also there are terms like Unet not mentioned in the main text}
}
\label{fig:figure_2}
\end{figure*}

\section{\Ours{}}
% In Section \ref{sec_method:background}, we first present relevant background information on latent diffusion models, human body representation, and human mesh recovery. Then, in Section \ref{sec_method:data_gen_pipeline} we introduce our proposed data generation method \Ours{}. Lastly, in Section \ref{sec_method:finetuning}, we demonstrate a downstream application of \Ours{} for single-view human mesh recovery \SY{again, be precise here, you don't use it standalone for HMR, you use it to contribute to a part of HMR}.

\subsection{Background}
\label{sec_method:background}
\textbf{Latent diffusion models.} Diffusion models are deep generative models that generate samples from a desired distribution by learning to reverse a gradual noising process. The sampling process starts from noise sampled from a standard normal distribution, which are refined into a series of less-noisy latents that eventually lead to the desired generation. For more details, please refer to Dhariwal \etal \cite{dhariwal2021diffusion} and Ho \etal \cite{ho2020denoising}. Latent Diffusion uses a perceptual compression model, a variational autoencoder (VAE) that projects the data distribution into a latent space, where the conditional diffusion process operates. Previous works \cite{avrahami2022latent} have shown that editing in the latent space is faster than pixel space editing \cite{avrahami2022blended} and helps to avoid pixel-level artifacts. Our method uses two latent diffusion models under the hood, a text-to-image model where the denoising is conditioned on the text input, and a depth-to-image model where the depth map is used as additional conditioning.

\textbf{SMPL body model.} We use the Skinned Multi-Person Linear (SMPL) model \cite{loper2015smpl} to represent the 3D mesh of the human body. SMPL is a differentiable function $\mathcal{M}(\theta, \beta)$ that takes a pose parameter $\theta \in \mathbb{R}^{69}$ and shape parameter $\beta \in \mathbb{R}^{10}$, and returns the body mesh $\mathcal{M} \in \mathbb{R}^{6890 \times 3}$ with $6890$ vertices. The 3D joint locations $X \in \mathbb{R}^{k\times 3} = \mathcal{W} \mathcal{M}$ are regressed from the vertices, using a pre-trained linear regressor $\mathcal{W}$, where $k$ is the number of joints.

\subsection{Data generation process}
\label{sec_method:data_gen_pipeline}
\Ours{} consists of three main steps. As a first step, we leverage a text-to-image model (i.e. Stable Diffusion \cite{Rombach_2022_CVPR}) to produce an initial generation of a posed person. Second, we predict the pose of the person and determine the difficulty level of the pose in the initial generation using a pose prior. We observe that Stable Diffusion tends to generate worse anatomy on more difficult poses. Thus, if the initial generation contains a hard pose, we render the depth map of the predicted body mesh taking into consideration the occlusion from other objects in the image. The depth map serves as the human structure prior. In the final step, we use the context information (in the form of image latents) from initial generation as a starting point, and leverage a depth-to-image model (i.e. a fine-tuned version of Stable Diffusion) to produce final generations by conditioning on the depth map from previous step.

Figure \ref{fig:figure_2} shows an overview of our method.
% \SY{this whole subsection is too sparse. Give the reader the main ideas and concepts before jumping into the specifics of each step.} 
Concretely, given a text prompt $t$ that describes the action of a person, we first encode it to text embeddings $z_t$, and then
% \SY{per comment above, reader does not understand at this point why we are first doing this. Please this in the context of the overall idea first} 
use a text-to-image Stable Diffusion model ($\mathcal{G}$) to generate an image $\mathcal{I} = \mathcal{G}(z_t)$. The image is passed to encoder $\mathcal{E}_{\mathcal{G}_d}$ of the compression module of $\mathcal{G}_d$, a depth-to-image diffusion model, to get the compressed image latents $z \in \mathbb{R}^{4 \times 64 \times 64}$, i.e. $z = \mathcal{E}_{\mathcal{G}_d}(\mathcal{I})$.
% \begin{align}
    % z =& \mathcal{E}_{\mathcal{G}_d}(\mathcal{I})
% \end{align}
% \SY{again, reader doesn't have context as to why this is happening next} 
Image latents $z$ contain context information about $\mathcal{I}$ such as the texture of the image and background layout. We can use $z$ as a starting point in the final generation process so the context from the inital generation is roughly preserved.

Next, we use an off-the-shelf model to reconstruct the 3D mesh of the person in $\mathcal{I}$. Specifically, we estimate the human pose $\theta$, shape $\beta$, and parameters of a weak perspective camera $\Pi$ from the image using an off-the-shelf HMR model $f: \mathcal{I} \rightarrow (\theta, \beta, \Pi)$. Since our method focuses on rectifying implausible human generations, we determine whether the initial generation is likely to contain implausible humans and only apply our rectification process on images with hard poses. We observe that Stable Diffusion tend to generate worse anatomy on more difficult poses. Hence, we use a pre-trained human pose prior VPoser \cite{SMPL-X:2019} as a proxy for determining if the person in image $\mathcal{I}$ has a challenging pose.
% as challenging poses tend to have large variance in the VPoser embedding space.
% \textbf{Use VPoser \cite{SMPL-X:2019} to determine pose difficulty.}
VPoser is a Variational Auto-Encoder (VAE) that is trained on a massive database of realistic human poses \cite{mahmood2019amass}. By design, poses that are farther away from the canonical pose (i.e. challenging poses) have larger variance in the embedding space. Therefore, we identify a difficult pose $\theta$ if its embedding $e_{\theta}$ have larger norm, i.e. $||e_{\theta}||_2 > \tau$, where $\{\mu, \sigma\} = \mathcal{E}_{v}(\theta)$ and $e_{\theta} \sim \mathcal{N} (\mu, \sigma)$. 
$\mathcal{E}_{v}$ is the encoder of VPoser. $\tau$ is determined empirically and set to 30.

% If the estimated pose is not challenging enough we sample a new image. \SY{confusing / not sure what is going on... Unclear why you would want to sample a new image if the pose is not challenging enough. Isn't it fine to leave it as is? Needs explanation} This process encourages variety and difficult examples in our generated images. \SY{hi Jen explain, so far you have said you just want to fix SD, not necessarily increase variety and difficulty}
% \SY{I will stop making this comment, you can apply to everything else, but again not clear to the reader what the point of attaining the silhouette is. Please add much more intuition about each of your ideas, as well as why they are interesting, solve specific challenges, or novel} 

Now that we have the predicted human pose from $\mathcal{I}$, we move on to the final step of our method where we inject pose information $\mathcal{M}(\theta, \beta)$ into the generation process to produce a more plausible image of a person with the predicted pose $\theta$. We achieve this by leveraging a depth-to-image version of Stable Diffusion, and using the depth values of the predicted human body as conditioning information in the generation process. Specifically, we render the 3D mesh to obtain the depth map $d_{fg} \in \mathbb{R}^{64 \times 64}$. Since there might be other objects in the image that occlude part of the person, we use Mask R-CNN \cite{he2017mask} (pre-trained on COCO \cite{lin2014microsoft}) to segment the non-human objects in the image and use the segmentation masks to mask out the occluded body part in the depth map $d_{fg}$. Formally,
% attain the person's silhouette $s$ from the positive depth values of the depth rendered person. Aided by the silhouette, we update the depth map with occlusions in the scene by using the object masks $m$ from a pre-trained segmentation model \SY{explain the segmentation model more. What objects are considered?}.%Both $m$ and $s$ are downsampled to the same dimension as the latents $z$.%We then use the union of $m$ and $s$ to select the background latents $z_{bg}$ we want to keep.
\begin{align}
    \{\theta, \beta, \Pi\} &= f(\mathcal{I}) \label{eq:mesh} \\
    % \{d_{fg}, s\} &= \mathcal{R}_d (\Pi, \mathcal{M} (\theta, \beta)) \\
    d_{fg} &= \mathcal{R}_d (\Pi, \mathcal{M} (\theta, \beta))  \label{eq:render_depth} \\ 
    d_{fg}^* &= d_{fg} \odot ((1 - m) \cap (d_{fg} > 0))
    %z_{bg} &= z \odot (1 - m \cup s) + \mathcal{N}(0, I) \odot (m \cup s)
\end{align}
where $\mathcal{R}_d$ is a depth renderer that renders the depth map of a mesh, $\odot$ denotes the Hadamard product, and $(d_{fg} > 0)$ is the silhouette of the rendered person.

Finally, to preserve the context information (e.g. texture and background layout) of the initial generation, we use initial image latents as a starting point in the final generation process. We add noise to $z$, and use a pre-trained denoising model (i.e. depth-to-image Stable Diffusion) to perform sequential denoising steps which produces the final image latents $z^*$. The denoising process (achieved through a pretrained UNet \cite{ronneberger2015u}) is guided by both the depth map $d_{fg}^*$ and text embeddings $z_t$. Final generation is obtained by decoding the $z^*$ with the compression module's decoder $\mathcal{D}_{\mathcal{G}_d}$. 
\begin{align}
    z^{noised} & = noise(z) \\
    z^* & = denoise(z^{noised}; d_{fg}^*, z_t) \\
    \mathcal{I^*} & = \mathcal{D}_{\mathcal{G}_d} (z^*) \label{eq:final_image}
\end{align}
As shown in Figure \ref{fig:figure_2}, final generation $\mathcal{I}^*$ contains similar texture and background as the original image, but the human body anatomy is rectified.

\subsection{Finetuning Human Mesh Recovery on challenging domains using synthetic data}
\label{sec_method:finetuning}
% \SY{need much more intro. Talk about how synthetic data can be useful for downstream tasks} 
Training a single-view Human Mesh Recovery (HMR) model end-to-end would require large amounts of images with paired 3D ground truths. Collecting such training sets requires burdensome motion capturing systems and is often limited to indoor laboratories. As a result, previous works such as \cite{weng2022domain} have focused on finetuning HMR model to a particular challenging domain using weak supervision (image paired with 2D keypoints). In this section, we introduce how  image-mesh pairs from \Ours{} can be used to finetuning HMR models in challenging domains.

Given a pre-trained HMR model that predicts pose $\theta$, shape $\beta$ and camera matrix $\Pi$ from an image $\mathcal{I}$ (i.e.
% \SY{model for what task? Also remind reader of variables since this is a new context / pretty different section}
$f: \mathcal{I} \rightarrow (\theta, \beta, \Pi)$), we aim to 
% that is pre-trained on standard Motion Capture (MoCap) and in-the-wild 2D pose estimation datasets \SY{is it both necessarily? This paragraph of broccoli crowns in a lot of information, take your time to set up more slowly}, and 
adapt the model to a new target-domain by finetuning $f$ on a small set of target images.
% \SY{also clarify $\mathcal{I}$ in this context}

% \SY{again, set up this task better. What is the precise setting, e.g. weakly supervised HMR w/ 2D ground truth, etc.}
In a typical finetuning setup where only 2D keypoints from the target are available as supervision, 2D reprojection loss can be minimized to encourage the consistency between predicted and ground truth keypoints. Formally, for an image from the target training set, let the ground truth 2D keypoints be $j \in \mathbb{R}^{k\times 2}$ with $k$ annotated keypoints per person, we would want to minimize $\mathcal{L}_{2D}^{real} = || \hat{j} - j||_2$
% \SY{remind / define projection matrices} 
% \begin{align}
%    \mathcal{L}_{2D}^{real} &= || \hat{j} - j||_2 
% \end{align}
where $\hat{j} = \Pi(\mathcal{W} \mathcal{M}(\hat{\theta}, \hat{\beta}))$ are the predicted 2D keypoints. Recall that $\mathcal{W}$ is the SMPL joint regressor, and $\Pi$ is the projection matrix of a weak perspective camera.

% \SY{rewrite better to highlight your key idea, not just run through description of what you do. E.g., given this task setting, Diffusion-HPC can be used to generate synthetic data that has xyz properties and offers XYZ benefits.} 
Given this task setting, \Ours{} can be used to generate synthetic data that has image-mesh pairs ($\mathcal{I}^*$ and $\{\theta, \beta, \Pi\}$ in Equations \ref{eq:mesh} and \ref{eq:final_image}). Then, on those synthetic image-mesh pairs, we can supervise the model with ground truth body parameters, which provide stronger form of supervision as compared to 2D keypoints.
\begin{align}
   \mathcal{L}_{3D}^{syn} &= || \hat{\beta} - \beta ||_2 + || \hat{\theta} - \theta ||_2 
\end{align}
Overall, loss during finetuning is $\mathcal{L} = \mathcal{L}_{2D}^{real} + \mathcal{L}_{3D}^{syn}$.
% \begin{align}
%     \mathcal{L} &= \mathcal{L}_{2D}^{real} + \mathcal{L}_{3D}^{syn}
% \end{align}

\textbf{Guidance from real images.} 
% \SY{don't get into empirical results yet. Instead say something like, in the case of a clear target data domain for the downstream HMR task, it can be useful to produce... also ``real guidance'' is kind of unclear. Maybe guidance from real images? or something better} Empirically we observe that it is better to produce synthetic images generated with real images as guidance. 
In the case of a clear target data domain for the HMR task, it can be useful to produce training data that have similar appearances to the target training set. Specifically, instead of using T2I diffusion model to generate the initial $\mathcal{I}$, we use real images from the training set. This guidance helps reduce the domain gap between the generated and real images because the appearances and poses will be more similar to the expected ones. 

\textbf{Pose augmentation.} 
% \SY{smoother connection. e.g., Finally, we can further enhance the diversity... Also explain a little more, you slap a bunch of equations without really explaining in English what is going on} 
Finally, we can further enhance the diversity of the generated poses, by applying pose augmentations to the predicted poses. Specifically, after Equation \ref{eq:mesh}, we can augment $\theta$ before proceeding to Equation \ref{eq:render_depth}. Formally, we apply pose augmentation in the embedding space of VPoser as in \citet{weng2022domain},
% Furthermore, to enhance the diversity of the generated poses, we apply augmentations to the poses. We do this in the VPoser embedding space of $\theta$ as in \citet{weng2022domain}. Formally, 
\begin{align}
    \mu, \sigma = & \mathcal{E}_v(\theta) \\
    z_{\theta}^{aug} =& z_{\theta} \odot (1 + s \epsilon), z_{\theta} \sim \mathcal{N}(\mu, \sigma) \\
    \theta^{aug} =& \mathcal{D}_v (z_{\theta}^{aug})
\end{align}
where $\mathcal{E}_v$ and $\mathcal{D}_v$ are the encoder and decoder of VPoser, $s$ is a constant scalar, and $\epsilon$ is from a multivariate uniform distribution of the same dimension as the VPoser latent space.

\section{Experiments}
\label{sec:experiments}
We first show the effectiveness of \Ours{} for improving HMR performance in challenging domains in Sec.~\ref{subsec:hmr}. Then, in Sec.~\ref{subsec:gen_quality} we present comparisons on the synthetic data generation quality of \Ours{}.

\subsection{Finetuning on challenging HMR settings}
\label{subsec:hmr}
We demonstrate the potential of \Ours{} through the task of few-shot adaptation of human mesh recovery models. We consider the setting where a small set of real images with 2D keypoints are available. This represents a typical scenario where we want to deploy a pre-trained HMR model on a new domain but there is limited ground truth annotations on the target domain. Through our experiments, we show that training with synthetic data from \Ours{} improves HMR on challenging target domains as compared to previous adaptation methods.

We use the following sports datasets as they contain much more challenging poses than common HMR benchmarks. As a result, there is a large domain gap when applying pre-trained HMR models on those datasets, and finetuning is necessarily to close the domain gap. Pre-processing details are in the \textit{Supplementary Material}.
% \SY{motivate choice of data sets} 

\begin{figure*}
\centering
\includegraphics[width=0.85\textwidth]{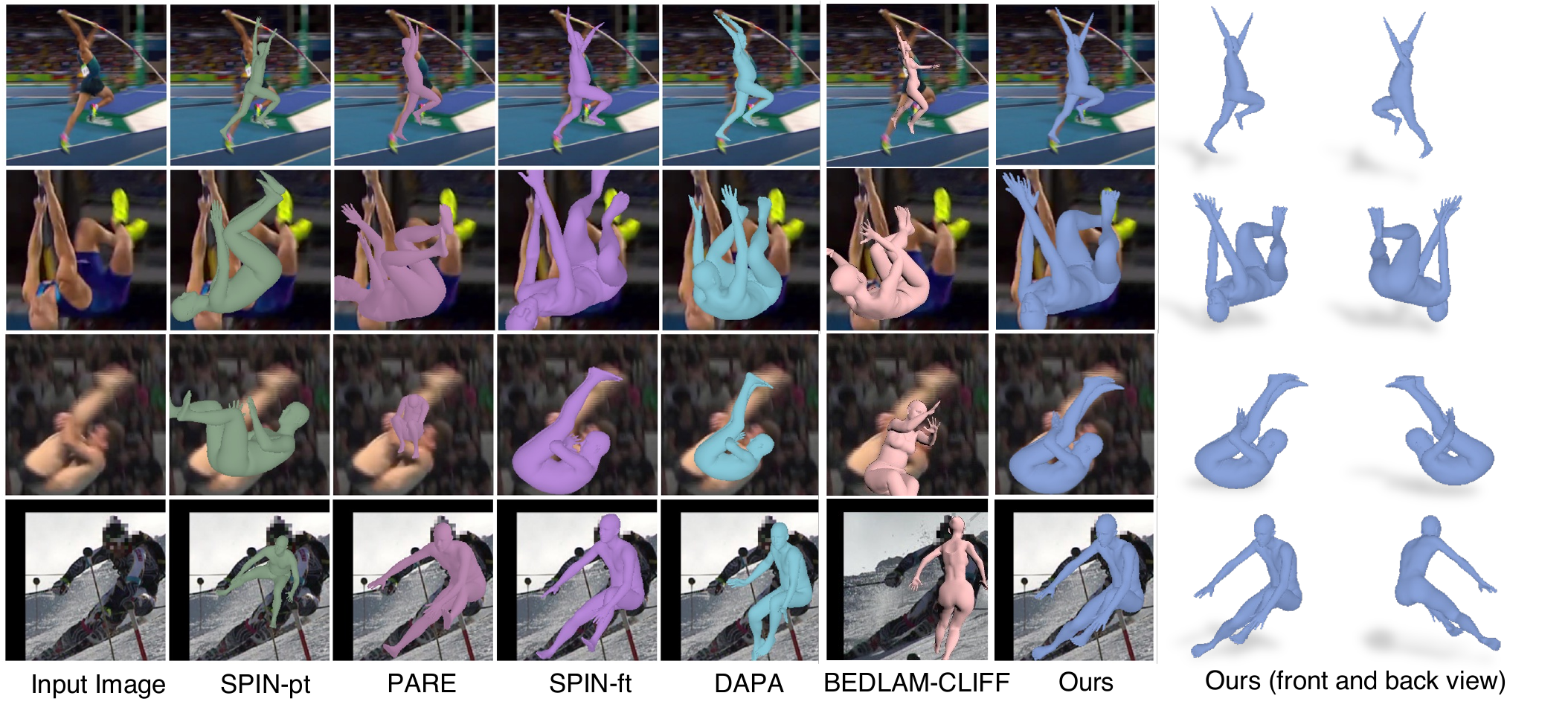}
\caption{Qualitative HMR results on SMART and Ski-Pose datasets. Finetuning with data from \Ours{} (rightmost) helps HMR models learn novel poses from challenging domains.}
\label{fig:qualitative}
\end{figure*}
\begin{itemize}[noitemsep,topsep=0pt]
    \item \textbf{Ski-Pose \citep{rhodin2018learning,sporri2016reasearch}} includes 3D and 2D keypoints labels from 5 professional ski athletes in motion. There is a significant domain gap between ski poses and poses from other human pose estimation datasets, therefore Ski-Pose has been used as a benchmark in evaluating pose domain adaptation \cite{gholami2022adaptpose}.
    \item \textbf{Sports Motion and Recognition Tasks (SMART) \citep{chen2021sportscap}} contains videos with per-frame 2D keypoints for various competitive sports. We consider 6 publicly released categories except for ``badminton", which only contains one clip.
    % namely ``diving" (81,121 images), ``pole vault" (1,208 images), ``high jump" (787 images), ``vault" (242 images), ``balance beam" (1,001 images), ``uneven bars" (1,398 images). (We took out badminton because it only has one clip.) We also downsampled diving.
    We sample enough clips so that the training set contains roughly $100$ images per category, and evaluate our finetuned models on the remaining images.
\end{itemize}
\textbf{Evaluation metrics.} For Ski-Pose, we use Mean Per Joint Position Error (MPJPE) and Procrustes-Aligned MPJPE (PA-MPJPE) as our evaluation metrics. PA-MPJPE measures MPJPE after performing Procrustes alignment of the predicted and ground truth keypoints. SMART does not have ground truth 3D keypoints, so we report Percentage of Correct Keypoint (PCK) determined by distance between predicted and ground truth keypoints in pixels.

% \begin{figure}
% \centering
% \includegraphics[width=\columnwidth]{author-kit-3DV2024-v1.2/figures/rg_examples.pdf}
% \caption{Example synthetic images generated using real images as guidance during HMR finetuning. Each real image (left) is accompanied by 3 synthetic generations (right), each with a slightly augmented pose. \Jen{Can perhaps move to supplementary if there is not enough space.}}
% \label{fig:rg_examples}
% \end{figure}

\begin{table*}[h!]
\centering
% \scriptsize
\resizebox{0.8\textwidth}{!}{
\begin{tabular}[t]{@{} ll *8c @{}}
\toprule
% \multicolumn{1}{c}{Method} & ft. & PA-MPJPE & MPJPE  \\ 
% \midrule
% SPIN-pt &  &  &  \\
% BEV & & & \\
% \midrule
\multirow{2}{*}{Method} & \multirow{2}{*}{Ft.} & \multirow{2}{*}{Ft. with \textit{syn}} & \multicolumn{6}{c}{PCK ($\uparrow$) per Action} & \multirow{2}{*}{Mean} \\
 \cmidrule(){4-9}
 & &  & Diving & Pole Vault & High Jump & Uneven Bars & Balance Beam & Vault &  \\
 \midrule
SPIN \cite{rhodin2018learning} & \xmark & - & 63.1 & 60.0 & 73.3 & 36.2 & 74.2 & 61.4 & 50.4 \\
BEV \cite{sun2022putting} & \xmark & - & 55.9 & 52.5 & 68.8 & 12.9 & 62.0 & 38.9 & 48.5 \\
PARE \cite{kocabas2021pare} & \xmark & - & 63.3 & 65.2 & 77.9 & 31.8 & 71.2 & 53.9 & 60.5  \\
BEDLAM-CLIFF \cite{Black_CVPR_2023} & \xmark & - & 30.4 & 57.3 & 67.4 & 31.1 & 55.7 & 48.3 & 48.4 \\
\midrule
% SPIN-ft & \cmark & None & & & & & &\\
% DAPA & \cmark & None &  & & &  & &\\
% Ours & \cmark & None &  & & &  & &\\
% \midrule
SPIN-ft \cite{rhodin2018learning} & \cmark & \xmark & 74.3 & 73.5 & 78.1 & 41.9 & 84.1 & 64.3 & 69.3 \\
DAPA \cite{weng2022domain} & \cmark & \cmark & 70.9 & 64.8 & \textbf{79.4} & 42.0 & 79.4 & 64.5 & 66.8 \\
ControlNet \cite{zhang2023adding} & \cmark & \cmark & 70.6 & 65.3 & 74.2 & 43.4 & 83.6 & 62.3 & 66.6 \\
Ours & \cmark & \cmark & \textbf{79.2} & \textbf{77.7} & 78.1 & \textbf{44.1} & \textbf{85.1} & \textbf{66.9} &  \textbf{71.9}  \\
\bottomrule
\end{tabular}
}
\caption{Quantitative results (PCK) on SMART. Ft indicates fine 
tuning on test data. Best numbers are in \textbf{bold}.}
\label{table:downstream_sportscap}

\end{table*}
\begin{table*}[h!]
\centering
\scriptsize
\resizebox{0.8\textwidth}{!}{
\begin{tabular}[t]{@{} ll *6c @{}}
\toprule
 \multirow{2}{*}{Method} & \multirow{2}{*}{Ft.} & \multirow{2}{*}{Ft. with \textit{syn}} & \multicolumn{4}{c}{MPJPE ($\downarrow$) / PA-MPJPE ($\downarrow$)} \\
\cmidrule(){4-8}
 &  &  & 0$\%$ train & 1$\%$ train  & 5$\%$ train & 50$\%$ train & 100$\%$ train \\
 \midrule
 SPIN \cite{kolotouros2019learning} & \xmark & - & 225.1 / 120.2 &-&-&-&- \\
BEV \cite{sun2022putting} & \xmark & - & 313.5 / 125.1 &-&-&-&- \\
PARE \cite{kocabas2021pare} & \xmark & - & 234.9 / 113.6 & - & - & - & - \\
ProHMR \cite{kolotouros2021probabilistic} & \xmark & - & \textbf{122.7} / \textbf{ 
82.6}$^*$ &-&-&-&- \\
BEDLAM-CLIFF \cite{Black_CVPR_2023} & \xmark & - & 363.5 / 136.5 &-&-&-&- \\
\midrule 
% CanonPose (HPE) \cite{wandt2021canonpose} & \cmark & MV & - &-&-&-&-&-& 128.1 / 89.6 \\
% AdaptPose (HPE) \cite{gholami2022adaptpose} & \cmark & ? & - &-&-&-&-&-& 99.4 / 83.0 \\
% \midrule
SPIN-ft \cite{rhodin2018learning} & \cmark & \xmark & - & 206.9 / 115.6 & 161.3 / 103.6 & 127.8 / 91.7 & 133.7 / 92.3\\
DAPA \cite{weng2022domain} & \cmark & \cmark & - & 222.2 / 123.6 & 180.2 / 108.7 & 128.7 / 90.5 & 126.9 / 86.1\\
ControlNet \cite{zhang2023adding} & \cmark & \cmark & - & 194.6 / 106.2 & 144.1 / 94.0 & 118.2 / 85.3 & 114.2 / 83.4 \\
Ours & \cmark & \cmark  & - & \textbf{182.3} / \textbf{105.9} & \textbf{143.4} / \;\textbf{90.7} & \textbf{116.5} / \textbf{83.5} & \textbf{111.3} / \textbf{81.5} \\
\bottomrule
\end{tabular}
}
\caption{Quantitative comparisons on Ski-Pose. We report MPJPE/PA-MPJPE on the test set. (*: Note that ProHMR uses ground truth 2D keypoints for test-time optimization and therefore has an unfair advantage for this experiment. All other models (including Ours) perform direct inference on test set). The best number per configuration is in \textbf{bold}.}
\label{table:downstream_ski}

\end{table*}

\begin{figure*}[t]
    \centering
    \includegraphics[width=0.76\textwidth]{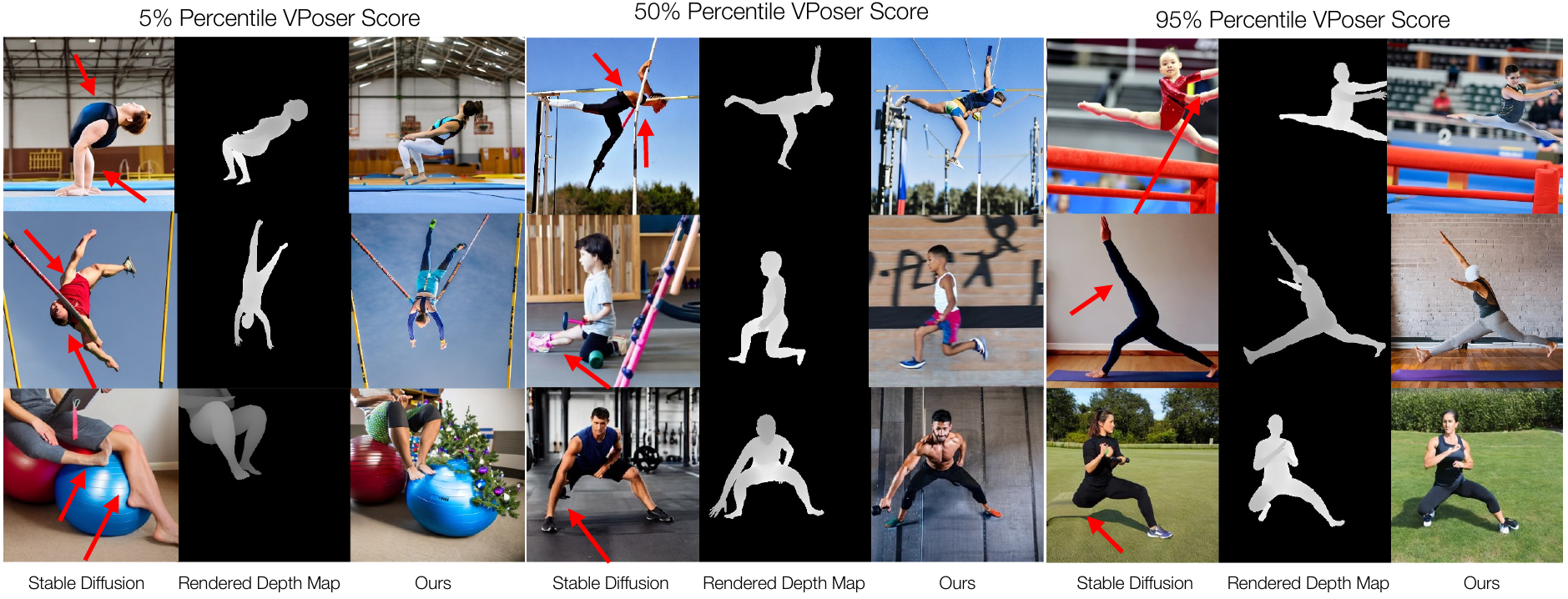}
    \caption{Comparison with Stable Diffusion \cite{Rombach_2022_CVPR} on text-conditioned generations. Red arrows point out implausible body parts in Stable Diffusion generations. To show a spectrum of varying pose difficulty levels, we present generations from the 5\%, 50\%, 95\% quantiles (i.e. from easy to hard) in terms of VPoser score. Rendered depths are included to show correct pose guidance.
    }
    \label{fig:generation}
\end{figure*}

\textbf{Implementation details.}
We use the backbone of SPIN \cite{kolotouros2019learning} to estimate the human mesh, since the backbone is shared by both SPIN and DAPA \cite{weng2022domain}, which enables fair comparison to these two. For each real image in the few-shot training set, we create $3$ synthetic images, where each one has a slightly different pose due to pose augmentation. We finetune and update the entire HMR model with batch size of 64, learning rate of 1e-4. All hyperparameters are the same as in SPIN. The models are trained until the loss curves plateau and on average each finetuning experiment takes about 6 hours on a single NVIDIA TITAN V GPU.

\textbf{Results.}
We compare to recent HMR models BEV \cite{sun2022putting} and PARE \cite{kocabas2021pare} that are pre-trained on MoCap datasets \cite{ionescu2013human3} as well as in-the-wild pose estimation datasets \cite{lin2014microsoft,andriluka20142d,johnson2010clustered}. We also compare to BEDLAM-CLIFF \cite{Black_CVPR_2023}, a state-of-the-art HMR model that is trained with a large synthetic dataset BEDLAM with realistic humans. In addition, we compare to finetuning methods SPIN-ft \cite{kolotouros2019learning}, DAPA \cite{weng2022domain}, as well as finetuning with synthetic data generated with ControlNet \cite{zhang2023adding} and \Ours{}. The finetuning of these methods and ours minimizes 2D keypoint reprojection error by using 2D keypoints from the target training set. In addition, SPIN-ft uses in-the-loop model fitting to provide additional model-based supervision. DAPA generates synthetic data with paired 3D ground truths on the fly as additional supervision, while Ours uses data from Diffusion-HPC.

In Table \ref{table:downstream_ski} we report PCK on sports categories from SMART. Although the off-the-shelf models (SPIN-pt, BEV, PARE) were pre-trained on 2D datasets that include sports poses \cite{johnson2010clustered}, there is still a significant domain gap between the training sets and SMART. BEDLAM-CLIFF was trained with a massive synthetic dataset BEDLAM, but the data generation was not tailored for the specific target domains, and therefore their training does not improve the model performance on the target dataset. As shown in the lower half of Table \ref{table:downstream_ski}, finetuning on a small set of target images is helpful in closing the domain gap. Among those methods, we achieve better performance in general.

In Table \ref{table:downstream_sportscap}, we report MPJPE/PA-MPJPE on Ski-Pose testset. We vary the size of the real training set during adaptation, and observe that with the same mount of real data, models trained with our synthetic data attain best performance. Further, with the help of synthetic data generated by \Ours{}, we attain better performance than SPIN-ft and DAPA using much smaller amount of real data. We achieve best performance when using the entire training set. Notably, our best performance (111.3 MPJPE, 81.5 PA-MPJPE) is better than ProHMR \cite{kolotouros2021probabilistic} (122.7 MPJPE, 82.6 PA-MPJPE), which uses ground truth 2D keypoints from the testset as additional information. Finally, as qualitatively demonstrated in Figure \ref{fig:qualitative}, our method produces more accurate human mesh estimations on challenging poses, and in general have better alignment with 2D images.

\subsection{Image generation quality}
\label{subsec:gen_quality}

\textbf{Data generation details.} We use a text-to-image Stable Diffusion \cite{Rombach_2022_CVPR} model pre-trained on LAION-5B \cite{schuhmann2022laion} and a CLIP ViT-L/14 \cite{radford2021learning} as text encoder. To condition the generation on depth maps we employ the depth-to-image Stable Diffusion model that was resumed from the text-to-image model, and finetuned for 200k steps. The denoising model has an extra input channel to process the (relative) depth prediction produced by MiDaS \cite{Ranftl2022} which is used as added conditioning. As our segmentation model we use Mask R-CNN \cite{he2017mask,wu2019detectron2} pre-trained on MS-COCO \citep{lin2014microsoft}. For the qualitative examples in Figure \ref{fig:generation} and experiments in Section \ref{subsec:gen_quality}, we use BEV \cite{sun2022putting} as the HMR model, due to its capacity of recovering people of all age groups and better empirical performance at localizing implausible synthetic humans, whereas two-stage HMR models that rely on a human detector often treat these erroneous generations as false negatives. With 50 inference steps, it takes about 6 seconds to create an image starting from text, and in the setting when a real image is used as guidance, the time is halved. 
% \vspace{-0.3cm}
\subsubsection{Comparison on text-conditioned generation}
We assess the quality of the text-only conditioned images generated by \Ours{} by comparing them to off-the-shelf Stable Diffusion. 
In order to span a wide taxonomy of human activities we compose text prompts from the category labels available in the MPII \cite{andriluka20142d} dataset.
In addition, to assess the generation quality regarding extremely challenging human poses, we use the publicly released  sports categories from SMART \cite{chen2021sportscap} (further introduced in Sec. \ref {subsec:hmr}) as text prompts. 
\begin{table}[ht]
\centering
% \begin{subtable}
\resizebox{\columnwidth}{!}{
\begin{tabular}[t]{@{} l *4c @{}}
\toprule
\multicolumn{1}{c}{Model} & Dataset & User Preference ($\uparrow$) & FID / H-FID ($\downarrow$) & KID / H-KID ($\downarrow$)\\ 
\midrule
Stable Diffusion & MPII & 0.45 $\pm$ 0.23 & \textbf{75.6} / 70.5 & \textbf{0.03} / 0.11 \\
\Ours & MPII & \textbf{0.55 $\pm$ 0.23} & \textbf{75.6} / \textbf{68.1} & \textbf{0.03} / \textbf{0.04} \\
\midrule
Stable Diffusion & SMART & 0.23 $\pm$ 0.08 & \textbf{66.3} / 91.4 & 0.04 / 0.07\\
\Ours & SMART & \textbf{0.77 $\pm$ 0.09} & 67.7 / \textbf{89.5} & \textbf{0.03} / \textbf{0.06} \\
\bottomrule
\end{tabular}
}
\caption{Text-conditioned comparisons on activities from MPII.
% Best numbers in \textbf{bold}.
}
\label{table:text_generation}
\end{table}

For both datasets, we report the standard evaluation metric Fréchet Inception Distance (FID) and Kernel Inception Distance (KID) \cite{heusel2017gans}. Since the focus of our method is on human generation, we report H-FID / H-KID, which is FID / KID computed with only foreground humans (segmented by Mask R-CNN). Note that FID/KID are computed using image-level features, and therefore do not focus on human generation quality in particular. Thus, we deem H-FID/H-KID more suitable metrics for our work.

Furthermore, we perform a user study where 6 independent blinded users were shown a randomly sampled set of 100 side-by-side images each generated by Stable Diffusion and \Ours{}. The users were given the task of selecting the image with the most plausible human pose and anatomy. If the images were comparable, the user could select a ``no preference" option.

\textbf{Results.} Table \ref{table:text_generation} presents comparisons on text-conditioned generations. While FID/KID values are roughly the same, we highlight that humans generated by \Ours{} have lower H-FID/H-KID to humans from real images. User study suggests that users prefer our generations most of the time. Qualitative results in Figure \ref{table:text_generation} suggest that our generations, while preserving the textures of the original images (hence similar FID/KID), effectively corrects the human anatomy (hence lower H-FID/H-KID). 
\vspace{-0.2cm}
\subsubsection{Comparison on pose-conditioned generation}
Most previous pose-conditioned generative models focus on the task of ``reposing" \citep{albahar2021pose,knoche2020reposing,men2020controllable}, where the goal is to repose the reference person using the target pose. These models are trained on fashion catalog images with clean background, therefore they are too simplistic to be effective baselines for our purpose. The only fair baseline, to our knowledge, is \citet{brooks2022hallucinating}, a recent StyleGAN2 \cite{karras2020analyzing}-based generative model that takes 2D keypoints of a posed person and generates images with compatible background. We benchmark their pre-trained model (trained on 18 million images sourced from 10 existing human pose estimation and action recognition datasets) on MPII for in-domain assessment.

% Since \Ours{} takes text and SMPL fittings as guidance, we choose to use MPII \citep{andriluka20142d}, a pose estimation dataset with taxonomy of every day human activities, as the evaluation set. We use action labels as text prompts, and SMPL fittings from EFT \citep{joo2020exemplar} as pose guide. 

\textbf{Results.} Table \ref{table:pose_generation} shows quantitative comparisons of image quality. 
% As our method is capable of generating an image with similar background to the real image (through the guidance of text and/or real image) while \citet{brooks2022hallucinating} generates arbitrary compatible backgrounds, we additionally report Human-FID (H-FID) for fair comparison of the foreground human generation quality. That is, FID computed using only the human region which we obtain by Mask R-CNN \cite{he2017mask}.  
% In addition, we report PCKh between the input and detected 2D keypoints as in previous work \cite{brooks2022hallucinating} to measure how well the image generation preserves the structure of the input skeleton. 
% Hallucinate Scenes has better PCKh, because it trains an explicit mapping from input pose to images, whereas \Ours{} infers the pose implicitly from text and rendered depths. 
Notably, even though \citet{brooks2022hallucinating} was trained with paired data (keypoint-image pairs) and therefore has an advantage over \Ours{} where the underlying models are trained/finetuned only with images, \Ours{} consistently achieves better performance. Moreover, \citet{brooks2022hallucinating} has poor generalization capability to novel pose distributions as in SMART, whereas our method powered by Stable Diffusion has a better zero-shot capability. Additional details, qualitative comparisons and limitations are in the \textit{Supplementary Material}.
\begin{table}[t]
% \vspace{-0.2cm}
% \scriptsize
\resizebox{\columnwidth}{!}{
\begin{tabular}[t]{lcccccc }  %p{0.1\textwidth}
\toprule
& \multicolumn{2}{c}{\textbf{Results on MPII}} & \\
\midrule 
\multirow{2}{*}{Method} & Trained with  & \multirow{2}{*}{T} & \multirow{2}{*}{R} & \multirow{2}{*}{D} & FID/& KID/ \\ 
& paired data & & & & H-FID & H-KID \\
\midrule 
Brooks et al.
 & \cmark & \xmark & \xmark  & \cmark & 109.0 / 75.5 & 0.10 / 0.07 \\ 
\midrule
\multirow{6}{*}{\Ours{}}
 & \xmark & \xmark & \cmark  & \xmark & 95.6 / 59.3 & 0.07 / 0.05 \\ 
 & \xmark & \cmark & \xmark  & \xmark & 54.6 / \underline{41.3} & 0.03 / 0.03 \\ 
 & \xmark & \cmark & \cmark  & \xmark & \underline{44.6} / 41.5 & \underline{0.02} / \underline{0.03}  \\ 
 & \xmark & \xmark & \cmark  & \cmark & 95.3 / 58.1 & 0.07 / 0.04 \\ 
 & \xmark & \cmark & \xmark  & \cmark & 72.8 / 136.2 & 0.03 / 0.11 \\ 
 & \xmark & \cmark & \cmark  & \cmark & \textbf{42.6} / \textbf{38.6} & \textbf{0.02} / \textbf{0.02} \\ 
\midrule
& \multicolumn{2}{c}{\textbf{Results on SMART}} & \\
\midrule 
Brooks et al.
 & \cmark & \xmark & \xmark  & \xmark & 175.8 / 114.1 & 0.14 / 0.06 \\ 
\midrule
\multirow{6}{*}{\Ours{}}
 & \xmark & \xmark & \cmark  & \xmark & 85.9 / 121.5 & 0.06 / 0.07 \\ 
 & \xmark & \cmark & \xmark  & \xmark & 162.3 / 113.9 & 0.13 / 0.08 \\ 
 & \xmark & \cmark & \cmark  & \xmark & 94.8 / \underline{99.8} & \underline{0.06} / \underline{0.05} \\ 
 & \xmark & \xmark & \cmark  & \cmark & \textbf{85.5} / 122.2 & 0.06 / 0.07 \\ 
 & \xmark & \cmark & \xmark  & \cmark & 145.9 / 131.5 & 0.10 / 0.07 \\ 
 & \xmark & \cmark & \cmark  & \cmark & \underline{92.4} / \textbf{44.5} & \textbf{0.06} / \textbf{0.03} \\ 
\bottomrule
\end{tabular}
}
\caption{Pose-conditioned generation quality. 
Note that \citet{brooks2022hallucinating} was trained with paired data (images with corresponding 2D keypoints), whereas \Ours{} is trained/finetuned only with images.
\Ours{} can take the background of text (``T") and/or a real image (``R") as conditioning information.}
% In-domain (``D") means the model had access to the dataset during training (for \cite{brooks2022hallucinating}) or finetuning (for \Ours{}). \textbf{Bold} and \underline{underlined} indicate best and next best numbers. 
\vspace{-0.4cm}
\label{table:pose_generation}
\end{table}
\vspace{-0.2cm}
\section{Conclusion}
\vspace{-0.2cm}
We proposed \Ours{}, a text-conditioned and training-free method that injects model-based human body prior to improve human-centric generation of state-of-the-art text-conditioned and pose-conditioned generative models. Further, \Ours{} demonstrates excellent utility in a challenging downstream task, single-view HMR.
For future work, we anticipate further investigation into the obstacles associated with human generation in foundation generative models, as well as exploring innovative ways of using generative models to tackle the challenges in 3D human perception tasks.

\paragraph{Acknowledgments.}
This material is based upon work supported by the National Science Foundation under Grant No. 2026498.
L. Bravo-Sánchez acknowledges financial support for this work by the Fulbright U.S. Student Program, which is sponsored by the U.S. Department of State and Fulbright Colombia. In addition, she was partially supported by an educational grant from IBM Research.
% This is used in the ARXIV version
\clearpage
\setcounter{page}{1}
\maketitlesupplementary
\renewcommand{\thefigure}{S\arabic{figure}} % LBS: added for clarity
% \section*{Supplementary Material}

\subsection*{A. Code}
% An anonymous version of the GitHub repository can be found at:
% \url{https://anonymous.4open.science/r/Diffusion_HPC-7887}
Code and trained models can be found at \url{https://github.com/ZZWENG/Diffusion_HPC}.

\subsection*{B. Additional Implementation Details}
\textbf{Quantitative evaluation in Table 1.} We compute quantitative metrics using roughly 10,000 images generated from MPII \cite{andriluka20142d} and SMART \cite{chen2021sportscap} prompts, respectively. 

For MPII, we use ``\{image description\} of \{person\} doing \{action\}" as the text prompts, where ``\{person\}" can be single- or multi-person descriptions of person(s) of interest, and ``\{action\}" are the activity categories from MPII. (We exclude categories ``inactivity, quite/light" and miscellaneous" because they do not describe a specific activity.) For example, resulting prompts could be ``a nice photo of a man doing water activities." or ``a high-resolution photo of a group of people doing conditioning exercises".

Text prompts for SMART are constructed using the template ``a photo of an athlete doing \{action\}" where action is one of ``high jump", ``vault", ``pole vault", ``diving", ``gymnastics on uneven bars", and ``gymnastics on a balance beam".

% \textbf{Use VPoser \cite{SMPL-X:2019} to determine pose difficulty.}
% VPoser is a Variational Auto-Encoder (VAE) that is trained on a massive database of realistic human poses \cite{mahmood2019amass}. By design, poses that are farther away from the canonical pose (i.e. challenging poses) have larger variance in the embedding space. Therefore, we identify a difficult pose $\theta$ if its embedding $e_{\theta}$ have larger norm, i.e. $||e_{\theta}||_2 > \tau$. 
% \begin{align}
%     \{\mu, \sigma\} &= \mathcal{E}_{v}(\theta) \\
%     e_{\theta} &\sim \mathcal{N} (\mu, \sigma)
% \end{align}
% $\mathcal{E}_{v}$ is the encoder of VPoser. $\tau$ is determined empirically and set to 30.

\textbf{Data processing for downstream experiments.}
Following previous works \cite{kolotouros2019learning,weng2022domain}, we crop the images such that the persons (localized by ground truth 2D keypoints) are centered in the crop. In addition, the persons are scaled such that the torso (i.e. mean distance between left/right shoulder and hip) are roughly one third of the crop size ($224 \times 224$).

\subsection*{C. Additional Comparisons on Pose-Conditioned Generation}

\begin{figure*}
    \centering
    \includegraphics[width=0.9\textwidth]{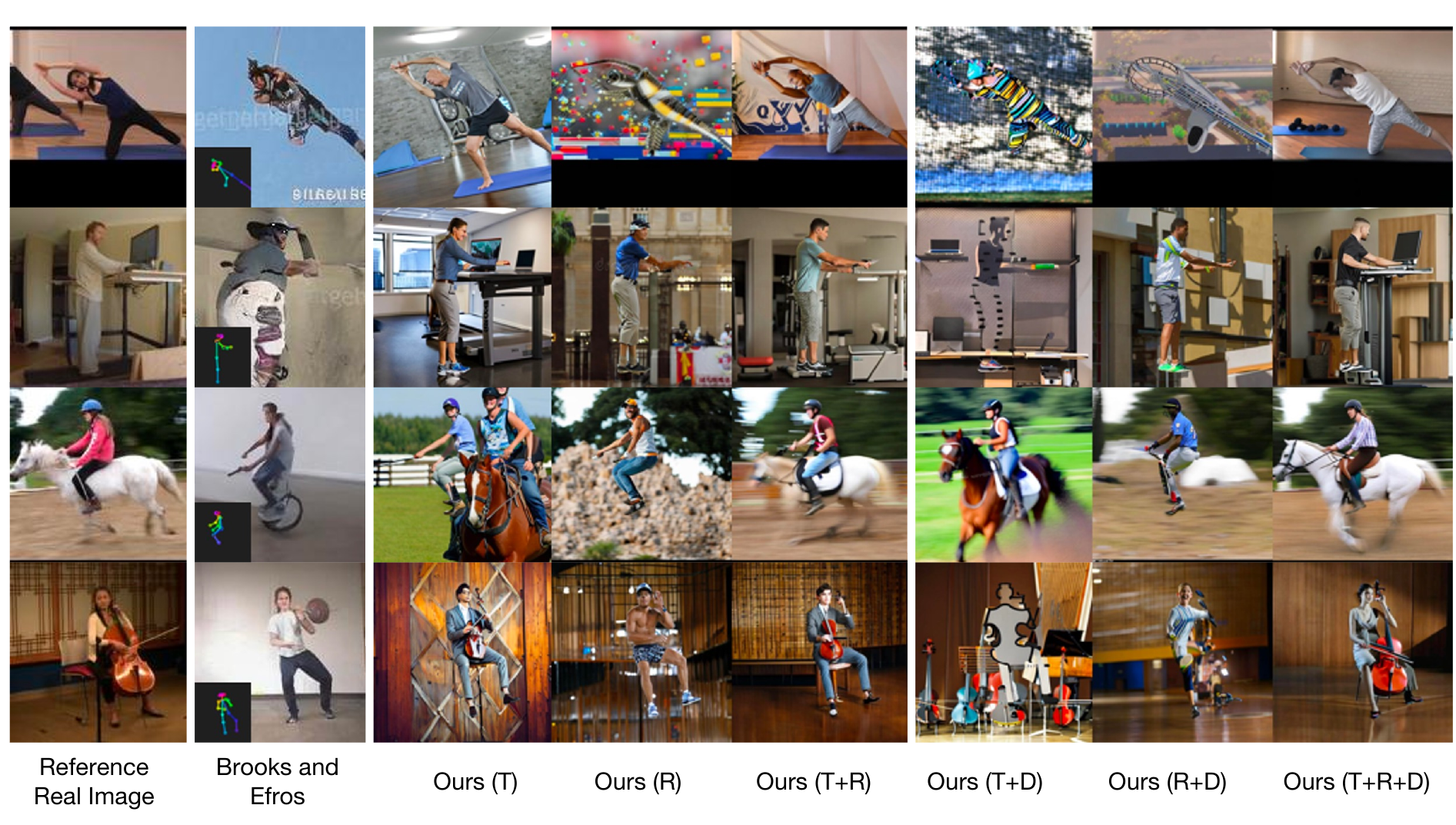}
    \caption{Qualitative comparisons to \citet{brooks2022hallucinating} (input 2D keypoints are overlaid on the bottom left). Our generations conditioned on text (T), real images (R), and in-domain (D).
    %.and both as guidance. In columns 2 to 4, we use off-the-shelf diffusion models, and in columns 5 to 8 we use finetuned diffusion models under the hood.
    % Best viewed by zooming in.
    % \SY{explain the different columns and summarize the takeaways more}
    % \vspace{-0.7cm}
    }
    \label{fig:hallucinate_comparison}
\end{figure*}

\textbf{Effect of text and real guidance.} Figure \ref{fig:hallucinate_comparison} demonstrates qualitative comparisons between different versions of our model and \citet{brooks2022hallucinating} As shown, text guidance (T) is essential in capturing the context of the human action. Guidance from real images (R) provides overall texture information such as background colors. While guidance from real images alone is not sufficient in preserving the action of the human ($3_{rd}$ row in Ours R), it adds to text guidance, and further improves the realism of the image generations (Ours T+R). 

\textbf{Effect of finetuning.} To see whether a finetuned diffusion model could further help improve generation quality, we finetune Stable Diffusion on the target dataset (MPII and SMART respectively) for 10 epochs. Generations with finetuned diffusion models is noted with ``D". As shown in Figure \ref{fig:hallucinate_comparison}, although finetuned diffusion model generates images with better background when there is no real guidance (Ours T vs. T+D), the foreground often loses the texture of humans, which is likely due to the ``catastrophic forgetting" as sometimes observed in finetuning large pretrained models. Qualitatively, with both text and real guidance, the effect of finetuning is barely noticeable (Ours T+R vs. T+R+D). Quantitatively, when using real guidance (with or without text guidance), finetuning slightly improves FID, and significantly improves H-FID and H-KID. Further, consistent to what is observed in qualitative results, 41.3 H-FID (with T) vs. 136.2 H-FID (with T+D) suggests that finetuning worsens performance without real guidance. This suggests that for text-conditioned generations, it is optimal to utilize an off-the-shelf diffusion model without finetuning.

\subsection*{D. Additional Qualitative Results}
Here we include additional qualitative results as well as failure cases for the text-conditioned and pose-conditioned generations (Section 4.2).
% \ref{subsec:gen_quality}).

\subsubsection*{Text-Conditioned Generation}
\begin{figure*}
    \centering
    \includegraphics[width=0.9\textwidth]{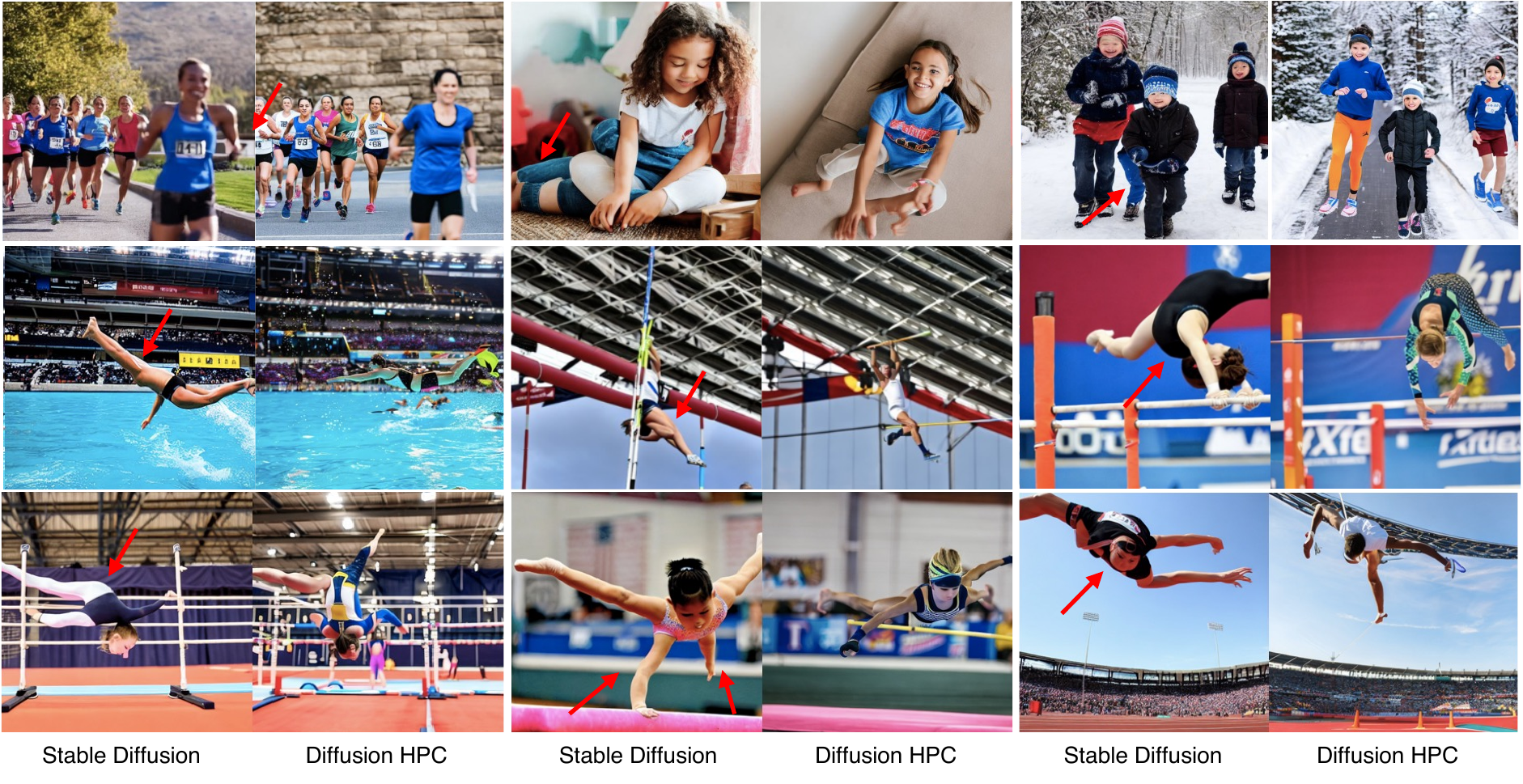}
    \caption{Comparison with Stable Diffusion \cite{Rombach_2022_CVPR} on text-conditioned generations. Row 1 and rows 2-3 are generated with MPII \cite{andriluka20142d} and SMART \cite{chen2021sportscap} prompts, respectively. Red arrows point out implausible body parts in Stable Diffusion generations.}
    \label{fig:qual_text}
\end{figure*}

Figure~\ref{fig:qual_text} shows qualitative comparisons of Stable Diffusion \citep{Rombach_2022_CVPR} and \Ours{} on text-conditioned generations. The images were selected from those sampled for the user study.

\begin{figure*}
    \centering
    \includegraphics[width=0.9\textwidth]{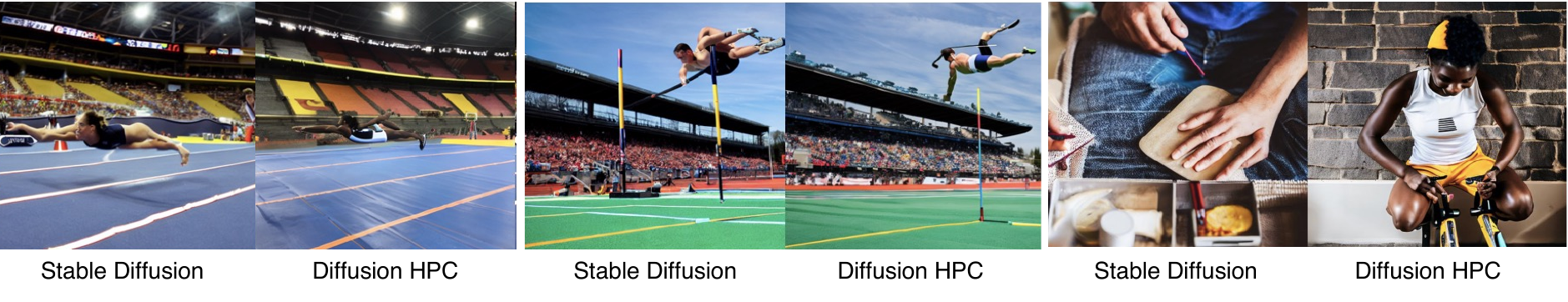}
    \caption{Failure cases on text-conditioned generations. }
    \label{fig:failure_text}
\end{figure*}

In Figure \ref{fig:failure_text} we include typical failure cases of text-conditioned generations. In left and middle columns, the body structures are not sufficiently rectified. This is likely because that resolution of depth maps (used for conditioning) is limited ($64\times 64$), so consequently small humans with out-of-distribution poses are challenging to rectify. In the right column, we show a failure scenario when the HMR model (i.e. BEV \cite{sun2022putting}) fails to reconstruct the humans in close-up shots. We could consider filtering out close-up shots, as they are not the primary intended use cases for \Ours{}.

\begin{figure*}
\centering
\includegraphics[width=0.95\textwidth]{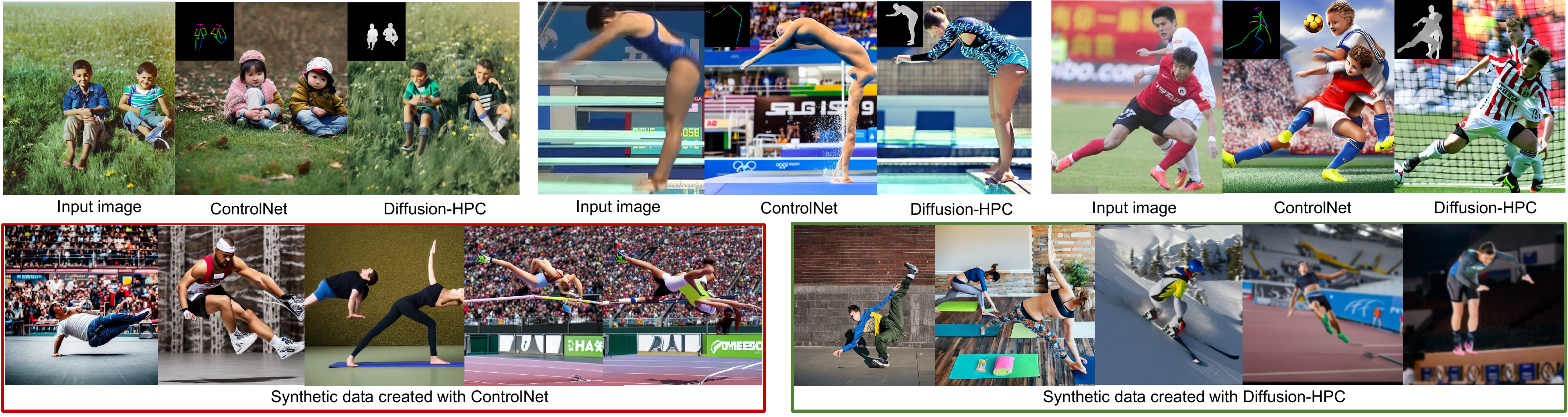}
\caption{Comparison to images generated by ControlNet. Top: Pose-to-image ControlNet often results in misaligned limbs (e.g. child legs in top left; front person's legs in top right), and does not generalize to challenging domains such as sports poses in general. Bottom: Example synthetic images generated by pose-conditioned ControlNet and Diffusion-HPC. We highlight Diffusion-HPC's superior generalization capability, even on extremely challenging domains such as sports. Such generalization capability is key for it to be effective for downstream tasks, such as HMR.}
\label{fig:controlnet}
\end{figure*}

\subsubsection*{Pose-Conditioned Generation}
\begin{figure*}
\centering
\includegraphics[width=0.9\linewidth]{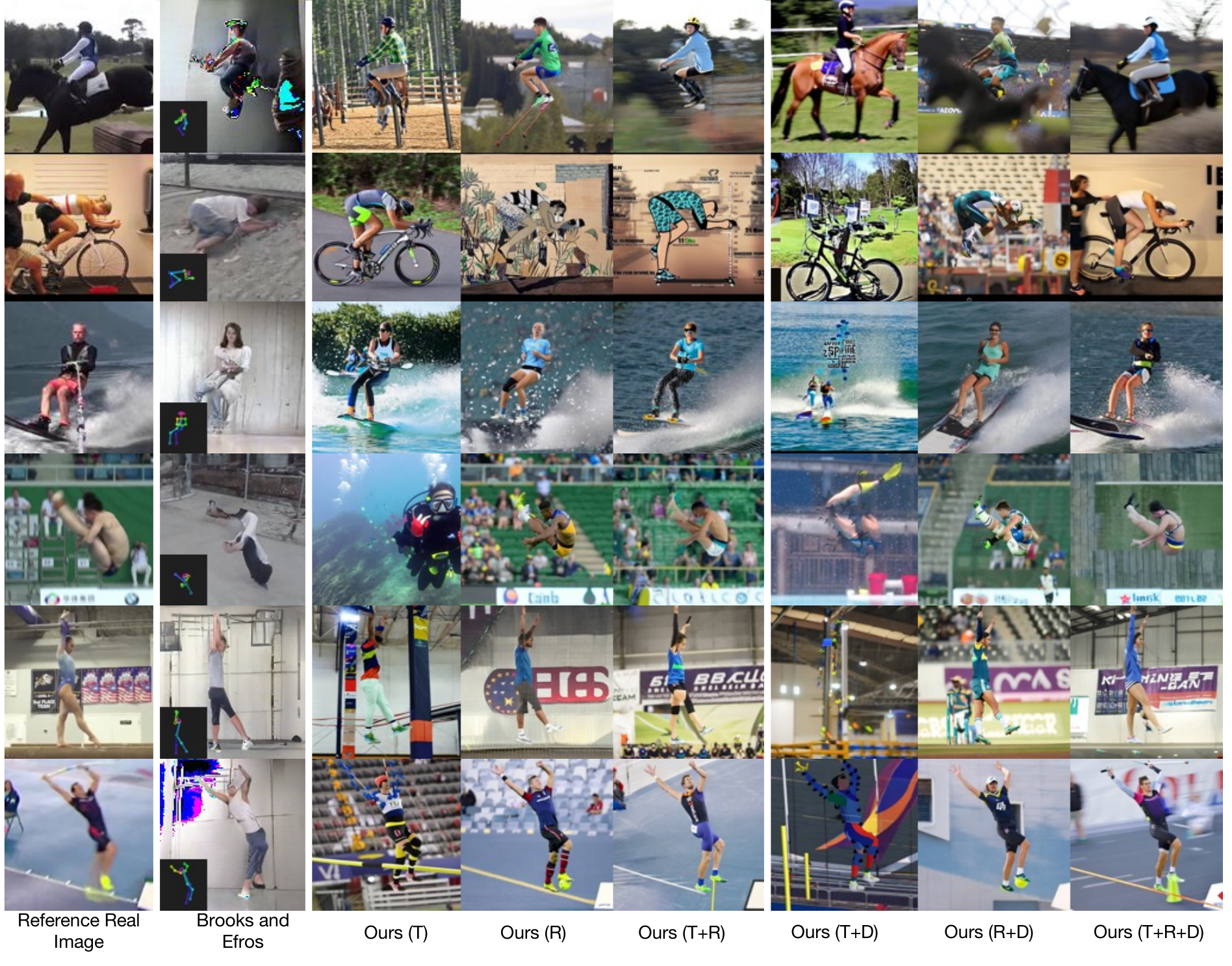}
\caption{Additional qualitative comparisons to \citet{brooks2022hallucinating} on the MPII dataset. Input 2D keypoints to \citet{brooks2022hallucinating} are overlayed on the bottom left in column 2. Top 3 rows are from MPII, and bottom 3 rows are from SMART. Our generations conditioned on text (T), real images (R). ``(D)" means the diffusion model is finetuned on the target dataset (MPII and SMART respectively).}
\label{fig:qual_pose}
\end{figure*}

\begin{figure*}
\centering
\includegraphics[width=0.9\linewidth]{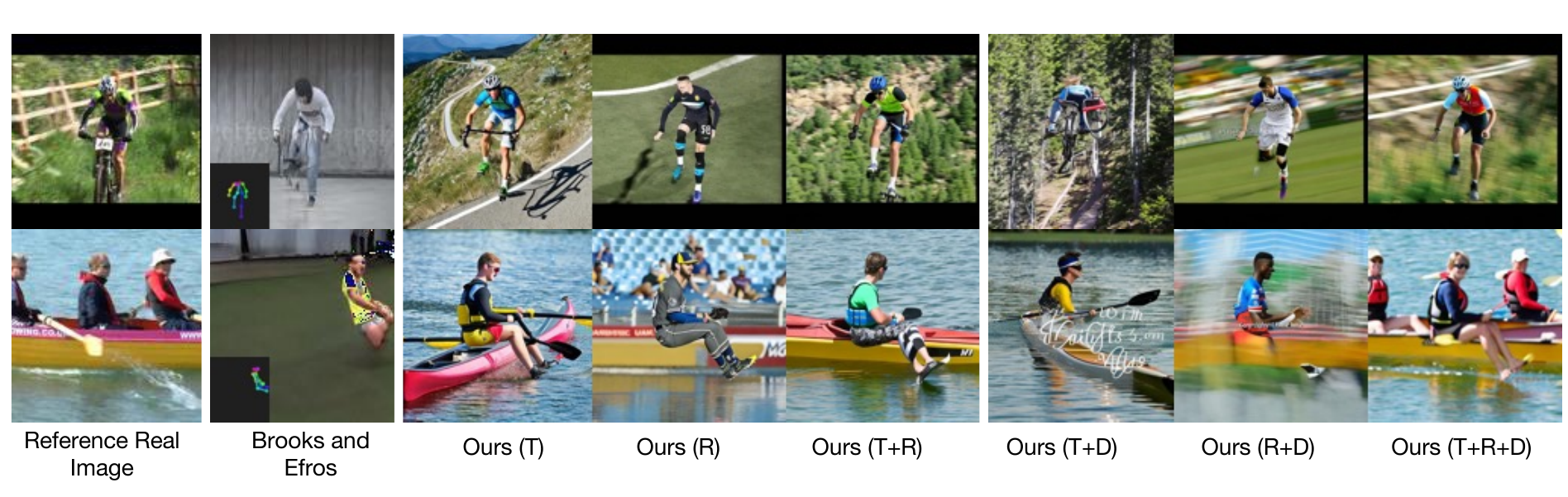}
\caption{Failure cases on pose-conditioned generations.}
\label{fig:failure_pose}
\end{figure*}

Figure~\ref{fig:qual_pose} shows qualitative comparisons of \citet{brooks2022hallucinating} and \Ours{} on pose-conditioned generations, and
Figure~\ref{fig:failure_pose} shows failures cases of pose-conditione generations. As seen from ``Ours T+R" and ``Ours T+R+D", human-object interactions are sometimes not preserved. 

Note that in \Ours{}, human-object interactions are considered but not modelled in an explicit way. Specifically, when we construct the depth map, we use Mask R-CNN \cite{he2017mask} to segment out the occluded body part, which helps with scenarios when, for instance, the person is riding the horse (row 1 of Figure \ref{fig:qual_pose}). However, row 2 of figure~\ref{fig:failure_pose} shows a failure case where the boat is not detected by Mask R-CNN.

In addition, in \Ours{}, latents from the initial generations help preserve the objects and context in the final generated scenes. For the pose-conditioned generations here, latents of real images help capture the background objects such as horse and surfboard (row 1 and 3 of Figure~\ref{fig:qual_pose}). However, when the background object is occluded or small (row 2 in Figure~\ref{fig:qual_pose} and row 1 in Figure~\ref{fig:failure_pose}), the latents are not sufficient in preserving the object in the final generations. Future work could consider extending \Ours{} by explicitly modelling the human-object/scene interaction. 

\textbf{Comparison to ControlNet \cite{zhang2023adding}.}
Adding control to pre-trained generative models has received increasing attention. While it is possible to use works such as ControlNet \cite{zhang2023adding} to generate pose-conditioned images, we note that ControlNet needs additional finetuning, requiring greater computing and annotation resources. In particular, training pose-to-image ControlNet requires paired data (i.e. 2D keypoints and images) which also limits the training data distribution to easy poses. 2D keypoints inherently provide less information compared to 3D body pose, and solely relying on them for 3D pose understanding tasks is insufficient. As a comparison, we train HMR models on SMART and SkiPose using the synthetic data generated with ControlNet, and on both evaluation sets, PCK and PA-MPJPE are worse than training with Diffusion-HPC (Table 1 and 2). Notably, for SMART, training with ControlNet data is worse than SPIN-ft which does not use synthetic data at all. Regardless of the possibility of re-training ControlNet using 3D human representations as conditioning, it is impractical when considering the substantial number of images with paired 3D GTs demanded. For reference, ControlNet pose-to-image model was trained on 200K keypoint-image pairs. Note that the scarcity of such 3D data was the primary motivation behind our work. Thus, rather than perceiving ControlNet as a direct alternative to our approach, it is more fair to consider it as a promising avenue to enhance our work, as the two methods can be synergistically combined - we could use Diffusion-HPC to bootstrap large amounts of image data with paired 3D pseudo ground truths and then use ControlNet to finetune a diffusion model.

\subsection*{E. Limitations}
As we rely on large pre-trained models \citep{Rombach_2022_CVPR,schuhmann2022laion}, any biases in these models or datasets that they were trained on will be replicated onto our generated images. Due to the resolution of depth maps ($64 \times 64$), fine details such as fingers and facial expressions are challenging to synthesize. Besides, since we only render person depth maps, human-object/human-scene interactions may not be well-preserved in the final generation (e.g. the person and yoga mat in column 3, row 2 of Figure 3).
%\ref{fig:generation}).
While these limitations do not affect downstream tasks where we only care about the body pose, there is large room to improve the photo-realism of human-centric image synthesis, and for the synthetic data to be useful for a wider variety of downstream tasks such as expressive HMR \cite{SMPL-X:2019} and recovering human-object/scene interaction \cite{bhatnagar2022behave,zhang2020perceiving,weng2021holistic}. Lastly, as we use SMPL body representation, our method does not consider people with limb losses, but it can be adapted to do so.

{
    % \newpage
    \small
    \bibliographystyle{ieeenat_fullname}
    \bibliography{main}
}

\end{document}